\pdfoutput=1

\documentclass[11pt]{article}

\usepackage{acl}

\usepackage{pgfplots}
\usepackage{times}
\usepackage{latexsym}
\usepackage[ruled,linesnumbered]{algorithm2e}
\usepackage{algorithmic}
\usepackage[T1]{fontenc}

\usepackage[utf8]{inputenc}
\usepackage{xcolor, soul}
\sethlcolor{yellow}
\usepackage{microtype}
\usepackage{float}
\usepackage{graphicx}
\graphicspath{ {./Images/} }
\usepackage{multirow}
\title{
TIMELINE: Exhaustive Annotation of Temporal Relations Supporting the Automatic Ordering of Events in News Articles
}

\author{Sarah Alsayyahi  \and Riza Batista-Navarro \\
         Department of Computer Science \\ University of Manchester  \\ \texttt{\{sarah.alsayyahi,riza.batista\}@manchester.ac.uk}}

\begin{document}
\maketitle
\begin{abstract}
Temporal relation extraction models have thus far been hindered by a number of issues in existing temporal relation-annotated news datasets, including: (1) low inter-annotator agreement due to the lack of specificity of their annotation guidelines in terms of what counts as a temporal relation; (2) the exclusion of long-distance relations within a given document (those spanning across different paragraphs); and (3) the exclusion of events that are not centred on verbs.
This paper aims to alleviate these issues by presenting a new annotation scheme that clearly defines the criteria based on which temporal relations should be annotated. 
Additionally, the scheme includes events even if they are not expressed as verbs (e.g., nominalised events). Furthermore, we propose a method for annotating all temporal relations---including long-distance ones---which automates the process, hence reducing time and manual effort on the part of annotators.
The result is a new dataset, the TIMELINE corpus, in which improved inter-annotator agreement was obtained, in comparison with previously reported temporal relation datasets.
We report the results of training and evaluating baseline temporal relation extraction models on the new corpus, and compare them with results obtained on the widely used MATRES corpus. 

\end{abstract}

\begin{table*}[h]
\centering
\begin{tabular}{|c|c|c|c|c|c|c|c|c|}
\hline
 & {\small Dataset} & {\small \# of documents} & {\small \# of relations} & {\small \begin{tabular}{@{}c@{}}Relation \\ window \end{tabular}} & {\small Type of events} & {\small IAA} \\
\hline
{\small 1} & {\small \begin{tabular}{@{}c@{}} TimeBank \\ \citep{pustejovsky2003timebank}  \end{tabular}}  & {\small 276} & {\small 6418} & {\small 0,1,..,n} & {\small \begin{tabular}{@{}c@{}}Events defined by \\  TimeML guidelines \end{tabular}} & {\small 55\%}  \\
\hline
{\small 2} & {\small \begin{tabular}{@{}c@{}} TimeEval3 \\ \citep{uzzaman2013semeval}  \end{tabular}} & {\small 276} & {\small 6418} & {\small 0,1,..,n} & {\small \begin{tabular}{@{}c@{}}Events defined by \\  TimeML guidelines \end{tabular}} & {\small 55\%}  \\
\hline
{\small 3} & {\small \begin{tabular}{@{}c@{}} TimeBank-DENSE \\ \citep{cassidy2014annotation}  \end{tabular}} & {\small 36} & {\small 12715} & {\small 0,1} & {\small \begin{tabular}{@{}c@{}}Events defined by \\  TimeML guidelines \end{tabular}} & {\small 64\%}  \\
\hline
{\small 4} & {\small \begin{tabular}{@{}c@{}} RED \\ \citep{o2016richer}  \end{tabular}} & {\small 95} & {\small 4969} & {\small 0,1} & {\small \begin{tabular}{@{}c@{}}Events defined by \\  Thyme-TimeML \\guidelines \end{tabular}} & {\small 41\%}  \\
\hline

{\small 5} & {\small \begin{tabular}{@{}c@{}} MATRES \\ \citep{ning2018multi}  \end{tabular}} & {\small 36} & {\small 1637} & {\small 0,1} & {\small \begin{tabular}{@{}c@{}}Events defined by \\  TimeML guidelines  \end{tabular}} & {\small 84\%}  \\
\hline
{\small 6} & {\small \begin{tabular}{@{}c@{}} TDDiscourse \\ \citep{naik2019tddiscourse}  \end{tabular}} & {\small 36} & {\small \begin{tabular}{@{}c@{}} 7x Larger\\ than TB-D \end{tabular}} & {\small 0,1,..,n} & {\small \begin{tabular}{@{}c@{}}Events defined by \\  TimeML guidelines  \end{tabular}} & {\small 69\%}  \\
\hline
\end{tabular}
\caption{\label{tab:datasets}
Comparison among most relevant temporal relation datasets in the literature.
}
\end{table*}
\section{Introduction}
Understanding the temporal structure of events in text is essential for a wide range of natural language processing tasks, e.g., question answering, information retrieval and inference \citep{campos2014survey,ng2014exploiting,ning2019cogcomptime}. Often, however, there is no explicit temporal information associated with most of the events in news articles. For instance, in the sentence \textit{``He \textbf{pointed} to the possibilities of new business models, products and ways of working that could have a dynamic impact on living standards.''}, there is no temporal expression associated with the event \emph{``pointed''} that conveys when exactly it occurred. 
The extraction of temporal relations, i.e., determining whether an event occurred before, after or at the same time as another event, makes it possible to capture the temporal sequence of events, even in cases where the text does not explicitly mention any temporal information with respect to an event \citep{ning2018multi, ning2019cogcomptime}.


Extracting temporal relations relies heavily on the annotation scheme adopted, which determines the granularity of the types of extracted temporal relations \citep{lim2019survey}.
In existing temporal information-annotated datasets \citep{pustejovsky2003timebank,bethard2007timelines,verhagen2007semeval,uzzaman2013semeval,cassidy2014annotation,reimers2016temporal,mostafazadeh2016caters,o2016richer,ning2018multi,naik2019tddiscourse}, many types of temporal relations are ignored, ill-defined or focussed only on specific types of events. In most datasets, only relations between events in the same or adjacent sentences are tagged \citep{naik2019tddiscourse}. Such limitation is the main reason for losing more precise temporal information for almost half of the events \citep{reimers2016temporal}. In addition, low agreement between human annotators is a common issue and needs to be improved by making the annotation task more clearly defined \citep{styler2014temporal,ning2018multi}. 
Our work seeks to address these issues by making the following contributions:
\begin{enumerate}
\item A novel annotation scheme with an unambiguous definition of the types of events and temporal relations of interest. We also provide a method for automatically identifying and annotating every possible temporal relation in a given document.
\item A new dataset called TIMELINE\footnote{Available at \url{https://github.com/Alsayyahi/TIMELINE}} consisting of 48 news articles, whereby a higher inter-annotator agreement was obtained in comparison with previously published temporal relation datasets.
\item An empirical analysis and an ablation study demonstrating the extent to which the TIMELINE dataset supports the development of models for ordering events in news articles.
\end{enumerate}


\section{Related work}

TimeBank is the first temporal information-labelled dataset to provide different types of temporal annotations (i.e., events, time expressions, and temporal relations) in news articles \citep{pustejovsky2003timebank}. However, there are two main issues with TimeBank: (1) the annotators tagged only temporal relations (referred to as TLINKs) which are considered as important \citep{pustejovsky2003timebank}, leading to sparse annotations; and (2) the scheme did not specify when two events should be paired up in a relation; as a result, inter-annotator agreement (IAA) was only around 55\%. Similar to TimeBank is the TimeEval3 dataset as it is a cleansed version of the former; it was created mainly for the TempEval shared tasks \citep{uzzaman2013semeval}. Meanwhile, the RED corpus considered different relations between events (e.g., temporal, coreference, causal and sub-event relations). It is a rich dataset created mainly to support the development of multi-task systems; the IAA for the relations of interest (e.g., ``before'' relations) is relatively low, i.e., around 41\% \citep{o2016richer}.

TimeBank-DENSE is a subset of TimeBank, which was introduced to address the sparsity issue in TimeBank by annotating all possible event pairs, and all event and time expression pairs in each given sentence and its surrounding sentences \citep{cassidy2014annotation}. However, many ill-defined temporal relations were annotated, leading to low IAA. The MATRES corpus tried to solve this issue by adopting a scheme that takes into consideration multiple timelines, i.e., \emph{axes}, distinguishing between events that actually happened (which belong to the main axis) and those which are only hypothetical (which belong to an axis parallel to the main one), for example. This multi-axis scheme required that each event relation is annotated while considering the relevant axis, thus improving the IAA significantly (84\%) \citep{ning2018multi}. However, they focussed only on events centred on verbs and ignored nominalised events.  

\newcite{cheng2018inducing} proposed automatically annotating temporal relations between events in a sentence and its surrounding sentences, using predefined rules based on the events' time anchors. They annotated temporal relations based on an existing dataset where the time anchors for events are already labelled \citep{reimers2016temporal}.  \newcite{naik2019tddiscourse} suggested a heuristic algorithm for the automatic inference of relations using the corpus developed by \newcite{reimers2016temporal}. Moreover, they made the first attempt to capture long-distance relations by asking experts to manually annotate a subset of unlabelled long-distance relations based on textual cues, external knowledge and narrative ordering. However, state-of-the-art models perform worse on their dataset, TDDiscourse, compared with other datasets. Error analysis shows that the models failed to deal with some of the phenomena in their dataset (e.g., negated/conditional events, event coreference, and the requirement to have access to real-world knowledge).

\begin{table}[h]
\centering
\begin{tabular}{|c|c|}
\hline
\textbf{{\small Event type}} & \textbf{{\small Category}} \\
\hline
 {\small Intension, Opinion  } & {\small  On an orthogonal axis  }\\
\hline
{\small Hypothesis, Generic  }  & {\small  On a parallel axis  }\\
\hline
 {\small Negation} & {\small  Not on any axis  }\\
\hline
{\small Static, Recurrent  }  & {\small Other}\\
\hline
\end{tabular}
\caption{\label{tab:axis}
Different event types that are not in the main axis.
}
\end{table}

\begin{figure*}[h]
 \centering
 \includegraphics[width=0.8\linewidth]{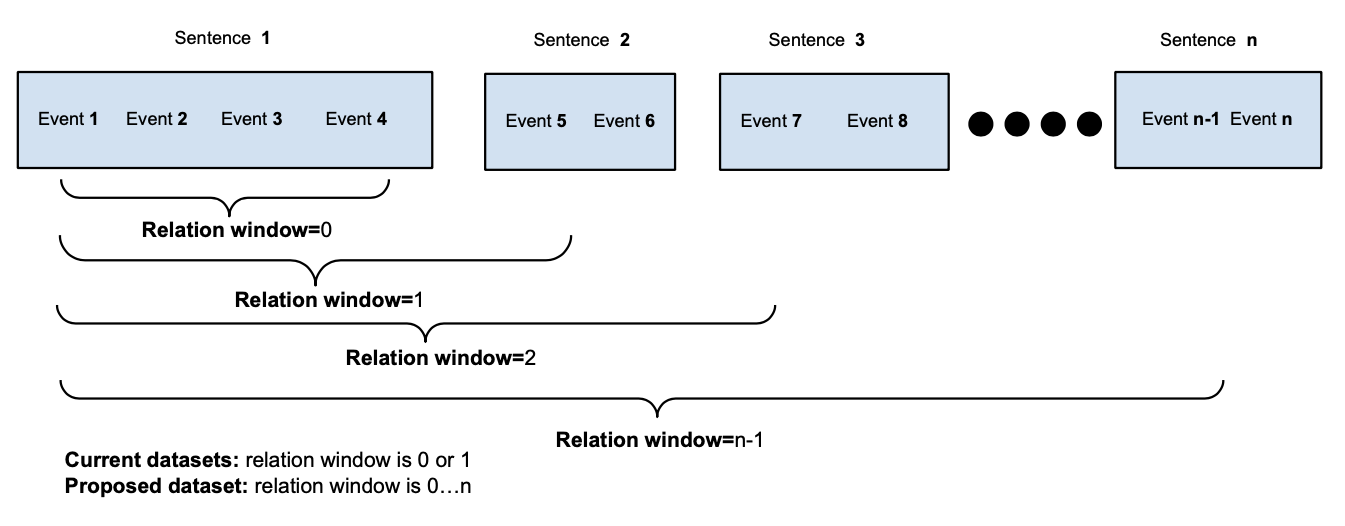}
 \caption{Illustration of the temporal relation window}
 \label{fig:Window}
\end{figure*}

\section{Motivation}
\label{sec:motivation}
This work is motivated by earlier relevant studies in the literature; we refer the reader to Table~\ref{tab:datasets} for more information about previously proposed temporal relation datasets. We, however, attempted to address the shortcomings of the previously proposed annotation schemes and developed a new dataset that specifies the relative order of events mentioned in a given news article. In designing our annotation scheme, we considered the following questions:
\begin{enumerate}
\item What types of events will be included? 
\item Is it possible to annotate the relations automatically based on the time anchors of events, and subsequently allow for retrieving the temporal order of any two events?  
\end{enumerate}

We reviewed existing temporal relation annotation schemes \citep{minard2015semeval,speranzanewsreader} to answer the first question. We then decided to discard events that cannot be anchored onto a timeline; these include intended, negated events and events involved in conditional constructions. Such events are the source of many ill-defined relations (e.g., vague relations) in existing datasets. A specific temporal relation between two events is labelled \emph{vague} if there is not enough information about the two events in the text that makes the annotator decide if the first event occurred before or after or at the same time as the second event. Consider the following example sentence: \emph{``She \textbf{planned} to \textbf{attend} the conference yesterday.''} The temporal relation between the two events (\emph{``planned''} and \emph{``attend''}) is vague as we cannot confidently determine the temporal relation between them  based on the context alone, i.e., it is possible that the event centred on \emph{``attend''} did not occur.

According to Ning et al., (2018), events belong to different time axes, hence the distinction between the following axes: (1) the main axis, i.e., a horizontal line where events that actually happened are represented (e.g., the event \emph{``planned''} in the example sentence); (2) an orthogonal axis (a vertical line that is orthogonal to the main axis) where opinions/intentions are placed (e.g., the event \emph{``attend''} in the same example); and (3) a parallel axis (a horizontal line parallel to the main axis) where generic and hypothetical events are placed. Hence, we focussed only on all events that belong to the main axis (events in the main storyline). We refer the reader to Table~\ref{tab:axis} for examples of events that do not belong to the main axis. We discuss details of how we identified events that need annotation in Section~\ref{subsection:annotation_scheme}.

Regarding the second question, we concluded that annotating every possible temporal relation in a specific news article is a non-feasible task \citep{naik2019tddiscourse}. Importantly, inconsistent temporal relation annotations are to be expected from a human annotator (e.g., a transitive constraint is not always satisfied) and have been noted in the TimeBank corpus \citep{bethard2007timelines,pustejovsky2011increasing}. Additionally, employing crowdsourcing for the annotation of these relations is expensive. For instance, \newcite{ning2018multi} reported that it costs about 400 USD to annotate temporal relations between events in a given sentence and its surrounding sentences in only 36 news articles. Also, \newcite{reimers2018event} highlighted that considering long-distance relations is required to retrieve correct temporal information for 40 \% of events in news articles. Therefore, to address these issues, we decided to automatically generate temporal relations and to directly infer consistent relations within different windows, i.e., relations between events which are separated by $0...n$ sentences. Further details on how temporal relations are generated will be given later in Section 4.2. Please refer to Figure~\ref{fig:Window} for an illustration of the relation window.   

\section{Dataset construction}
In this section, we describe the process for collecting the documents included in our corpus. This is followed by a discussion of the details of our proposed annotation scheme.

\subsection{Document collection}
The \textit{LexisNexis} library is an online resource that offers access to court cases, commentaries, handbooks and news articles, amongst others\footnote{\url{https://www.lexisnexis.com/uk/legal/}}. The library was used to retrieve a total of 48 news articles published in a UK newspaper: The Times (London). Table~\ref{tab:corpus} presents the queries that we used to retrieve the articles. 

\begin{table}[ht]
\centering
\begin{tabular}{|c|c|c|c|c|}
\hline
{\small No} & {\small Category} & {\small \# of articles} &  {\small \begin{tabular}{@{}c@{}}Publication \\ Period \end{tabular}}  \\
\hline
{\small 1} & {\small \begin{tabular}{@{}c@{}} ``Covid-19'' and \\ ``Economy''  \end{tabular}} & {\small 16} & {\small \multirow{2}{*}{\begin{tabular}{@{}c@{}}  \\  March 2020 - \\ Feburary 2021  \end{tabular}}}  \\
{\small 2} & {\small \begin{tabular}{@{}c@{}} ``Covid-19'' and \\ ``Vaccine''  \end{tabular}} & {\small 16} &     \\
{\small 3} & {\small \begin{tabular}{@{}c@{}} ``Covid-19'' and \\ ``Quarantine''  \end{tabular}} & {\small 16} &     \\
\hline
\end{tabular}
\caption{\label{tab:corpus}
Queries used in retrieving the news articles in the TIMELINE corpus.
}
\end{table}

\subsection{Annotation Scheme}
\label{subsection:annotation_scheme}
Our scheme consists of multiple layers of annotation which are described below.
\paragraph{Event annotation.} Events in our corpus were annotated according to the TimeML guidelines \citep{sauri2006timeml}, which define an \emph{event} as a situation that occurs. Events are centred on one or more trigger words and can be expressed in different ways. This includes verbs, e.g., \emph{``said''}, or phrasal verbs, e.g., \emph{``woke up''}, as well as nominal events, e.g., \emph{``World Cup''} or \emph{``demonstration''}. We included all events that can be anchored onto a timeline as long as they belong to the main axis. However, as discussed in Section~\ref{sec:motivation}, we excluded specific types of events: intended, negated, static, generic, and hypothetical events. In the Appendix, we provide a complete list of the broad types of events that we excluded, alongside some examples.

\paragraph{Time anchor annotation.} Drawing inspiration from previous studies, we adopted the use of the concept of narrative container (NC) in order to increase the accuracy of temporal relation annotation \citep{pustejovsky2011increasing}. NC is the default interval surrounding the document creation time (DCT) of an article, and provides an estimate of when a given event with no explicit time anchor, happened. It is affected by different variables related to text style and genre; for example, the NC value for newspapers is 24 hours, while that for weekly and monthly publications is a week and a month, respectively. 
Since our corpus consists of newspaper articles published on a daily basis, we can set the value of the narrative container to 24 hours---this was made clear to our annotators. Furthermore, annotators were provided with the DCT for every news article. Annotators were advised to use external and background knowledge if it helps them in providing more accurate time anchors. Where an event occurred over an interval, annotators were asked to provide the time anchor based on the start of the interval.

Earlier work which attempted to automatically generate temporal relations based on time anchors of events \citep{cheng2018inducing,naik2019tddiscourse} were hindered by their reliance on the EventTime corpus \citep{reimers2016temporal}. In this corpus, some events were given under-specified dates (e.g., \emph{``after 1990-XX-XX''}) which made it difficult to form temporal relation annotations involving such events. In contrast, in our annotation scheme, events are always given explicit or implicit dates.
Specifically, annotators were asked to enter the time anchor of the form \emph{YYYY-MM-DD} for each event, by choosing one of six possible options based on the type of temporal information associated with the event. For instance, if the temporal information associated with the event is mentioned explicitly in the text (e.g., \emph{``June 14, 2022''}), the annotator specifies \emph{``2022-06-14''}. If the temporal information associated with the event is mentioned in the text in a vague manner (e.g., \emph{``August''}), the annotator specifies \emph{``2022-08-XX''}. We refer the reader to the annotation guidelines in the Appendix for a list of all possible options with examples.

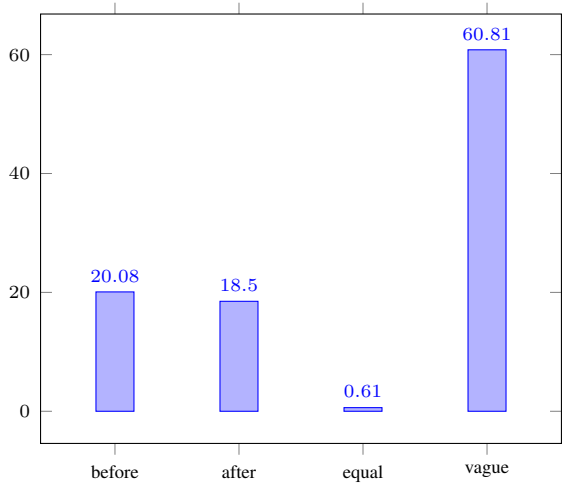
\begin{figure}[h]
\centering
    \begin{tikzpicture}[scale=1, every node/.style={scale=0.6},every node/.style={align=center,font=\scriptsize}] 
    \begin{axis}
    [title=, ybar, nodes near coords,bar width=0.5cm, symbolic x coords={before,after,equal,vague}, enlarge x limits=.2,]
    \addplot coordinates{(before,20.08)(after,18.50)(equal,0.61)(vague,60.81)};
    \end{axis}
    \end{tikzpicture}
\caption{Label distribution in the TIMELINE corpus.}
\label{fig:Dist}
\end{figure}
\paragraph{Temporal relation annotation.} Before the automatic generation of relation annotations, the annotators were asked to answer a set of questions for each annotated event. These questions, for example, help determine whether two events happened at the same time (and thus should be given the ``equal'' label), and help reduce the number of ``vague'' relations by prompting the annotator to consider details within the context of events.
We refer the reader to the annotation guidelines in the Appendix for a list of all the questions. Then, we developed a method for generating temporal relations that: (1) identifies every possible pair of events in a given document, and (2) generates consistent temporal relation labels based on the annotation given in the previous steps. The method handles every possible case to generate one of the following labels: \texttt{before},  \texttt{after}, \texttt{equal} and \texttt{vague} for each relation. For further details, we refer the reader to the Appendix, which shows the algorithm that generates a label for every possible relation. Figure~\ref{fig:Dist} shows the distribution of the generated temporal relation labels. As illustrated in the figure, most of the relations are ``vague'' due to the inherent ambiguity of temporal information in natural language text. 
\begin{table}[ht]
\centering
\begin{tabular}{|c|c|c|}
\hline
 \multicolumn{3}{|c|}{\textbf{{\small Inter-Annotator Agreement}}}  \\
\hline
{{\small Event annotation}} & \multicolumn{2}{|c|}{{\small Relation annotation}}  \\
\hline
{\small F1-score} & {\small Micro F1-score} & {\small Cohen's Kappa}\\
\hline
{\small 91.29} & {\small 92.98} & {\small 86.75} \\
\hline
\end{tabular}
\caption{\label{tab:iaa}
Inter-annotator agreement for event and temporal relation annotation.
}
\end{table}

\begin{table}[ht]
\centering
\begin{tabular}{|c|c|c|c|c|c|}
\hline
 & \textbf{{\small before}} & \textbf{{\small after}} & \textbf{{\small equal}} & \textbf{{\small vague}} & \textbf{{\small TOTAL}} \\
\hline
 \textbf{{\small before  }} & {\small  397  } & {\small  11  } & {\small  0  } & {\small  26  } & {\small  434  }\\
\hline
\textbf{{\small after  }} & {\small  8  } & {\small  336  } & {\small  1  } & {\small  28  } & {\small  373  }\\
\hline
\textbf{{\small equal  }} & {\small  2  } & {\small  0  } & {\small  10  } & {\small  2  } & {\small  14  }\\
\hline
\textbf{{\small vague  }} & {\small  36  } & {\small  17  } & {\small  0  } & {\small  1268  } & {\small  1321  }\\
\hline
\textbf{{\small TOTAL  }} & {\small  443  } & {\small  364  } & {\small  11  } & {\small  1324  } & {\small  2142  }\\
\hline
\end{tabular}
\caption{\label{tab:cm}
Contingency matrix for temporal relation annotation. Rows correspond to Annotator 1 while columns correspond to Annotator 2.
}
\end{table}

\section{Corpus Reliability}
Three annotators contributed to the annotation of our corpus: the first one (the first author of this paper) annotated all the articles, whilst the second and third annotators annotated 31\% of the articles.
Table~\ref{tab:iaa} presents the average inter-annotator agreement between the annotators at the level of events (calculated using F1-score) and temporal relations (calculated using micro-averaged F1-score and Cohen’s Kappa). It is worth noting that the agreement over temporal relation annotations is based on events that annotators agreed on.   

The contingency matrix in Table~\ref{tab:cm} shows the agreement and disagreement between the first and second annotators with respect to temporal relation annotation. One can observe that the agreement between the annotators is high for all temporal relation types, which implies that the annotation scheme led to consistent annotations. 

The second and third annotators are PhD Computer Science students who have received training on the proposed annotation scheme. Upon completion of the annotation tasks, they were compensated at an hourly rate of £15.

Three subsets were defined, containing randomly selected documents: training (70\%), development (10\%) and test (20\%). We refer the reader to Table~\ref{tab:overview} for details on the number of documents and event pairs (annotated with temporal relations) in each subset.

\begin{table}[h]
\centering
\setlength\tabcolsep{5pt}
\begin{tabular}{|c|c|c|}
\hline
\textbf{{\small Data splits}} & \textbf{{\small Number of documents}} & \textbf{{\small Number of pairs}} \\
\hline
\textbf{{\small Train}} & {\small 23} & {\small 2384} \\
\hline
\textbf{{\small Dev}} & {\small 11} & {\small 284} \\
\hline
\textbf{{\small Test}} & {\small 14} & {\small 685} \\
\hline
\end{tabular}
\caption{\label{tab:overview}
Number of documents and event pairs annotated with temporal relation types in the TIMELINE corpus. 
}
\end{table}
\begin{table}[h]
\centering
\setlength\tabcolsep{5pt}
\begin{tabular}{|c|c|c|c|c|c|c|}
\hline
\multirow{2}{*}{}&\multicolumn{3}{|c|}{\textbf{{\small Baseline method 1}}}&\multicolumn{3}{|c|}{\textbf{{\small Baseline method 2}}} \\ \cline{2-7}
 & \textbf{{\small P}} & \textbf{{\small R}} & \textbf{{\small F1}} & \textbf{{\small P}} & \textbf{{\small R}} & \textbf{{\small F1}}  \\
\hline
\multirow{4}{*}{}{\small before} & {\small 77.59} & {\small 86.89} & {\small 81.97} & {\small 79.03} & {\small 86.14} & {\small 82.43}  \\
{\small after} & {\small 78.57} & {\small 86.51} & {\small 82.35} & {\small 79.03} & {\small 86.14} & {\small 82.43}  \\
{\small equal} & {\small 0}& {\small 0} & {\small 0} & {\small 0} & {\small 0} & {\small 0}  \\

\hline
{{\small mi-F1}}  & \multicolumn{3}{|c|}{{\small 81.44}} & \multicolumn{3}{|c|}{\small 81.70} \\  
\hline
 
\end{tabular}
\caption{\label{tab:results1}
Performance of two baseline methods on the MATRES corpus (mi-F1 pertains to micro-averaged F1-score).
}
\end{table}

\begin{table}[h]
\centering
\setlength\tabcolsep{5pt}
\begin{tabular}{|c|c|c|c|c|c|c|}
\hline
\multirow{2}{*}{}&\multicolumn{3}{|c|}{\textbf{{\small Baseline method 1}}}&\multicolumn{3}{|c|}{\textbf{{\small Baseline method 2}}}\\ \cline{2-7}
 & \textbf{{\small P}} & \textbf{{\small R}} & \textbf{{\small F1}} & \textbf{{\small P}} & \textbf{{\small R}} & \textbf{{\small F1}}  \\
\hline
\multirow{4}{*}{}{\small before} & {\small 67.86} & {\small 67.25} & {\small 67.55} & {\small 69.05} & {\small 70.07} & {\small 69.55}  \\
{\small after} & {\small 67.14} & {\small 69.62} & {\small 68.36} & {\small 69.05} & {\small 70.07} & {\small 69.55}  \\
{\small equal} & {\small 100}& {\small 5} & {\small 9.52} & {\small 0} & {\small 0} & {\small 0}  \\
\hline
{{\small mi-F1}}  & \multicolumn{3}{|c|}{{\small 67.51}} & \multicolumn{3}{|c|}{{\small 69.05}} \\  
\hline
 
\end{tabular}
\caption{\label{tab:results2}
Performance of two baseline methods on the TIMELINE corpus (mi-F1 pertains to micro-averaged F1-score).
}
\end{table}
\section{Baseline methods}
In order to assess the extent to which our proposed TIMELINE corpus supports the development of temporal relation extraction approaches, we sought to train and evaluate two baseline models for temporal relation extraction.
Specifically, we employed two temporal relation classification models proposed by \newcite{han2019deep} as baseline methods. Both models are based on bidirectional long short-term memory networks (BiLSTMs), but one of them re-optimises the network to adjust for global properties, i.e., symmetry and transitivity constraints. These models were selected based on their highly competitive performance and the availability of their source code.


We note that prior to training and evaluating each of the said models, all temporal relations labelled as \texttt{vague} in the TIMELINE corpus, were discarded for the following reasons:
(1) this type serves as a catch-all category for any relations which are ambiguously expressed in text and yet is over-represented (accounting for 60.81\% of the annotated relations); and
(2) more importantly, the performance for the \texttt{vague} relation type was not considered in previously reported work---they treated this label similarly to how they handled events with no temporal relations between them \citep{ning2019improved}.

In preparation for training the models, we generated BERT embeddings \citep{devlin2018bert} and part-of-speech (POS) embeddings for every token in the sentences containing events that are involved in a temporal relation, as both models take these as input representation. For the training process, we adopted the hyperparameter values used by \newcite{han2019deep}.


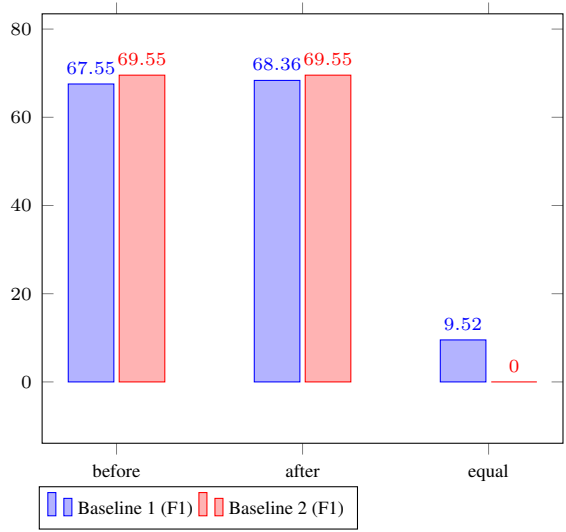
\begin{figure}
    \begin{tikzpicture}[scale=1, every node/.style={scale=0.6},every node/.style={align=center,font=\scriptsize}] 
    \centering
    \begin{axis}  
    [ybar, enlargelimits=0.20, legend style={at={(0.3,-0.1)}, anchor=north,legend columns=-1}, ylabel={}, symbolic x coords={before, after, equal}, enlarge x limits=.2, xtick=data, nodes near coords, bar width=0.6cm, nodes near coords align={vertical},]  
    \addplot coordinates {(before, 67.55) (after, 68.36) (equal, 9.52) };   
    \addplot coordinates {(before, 69.55) (after, 69.55) (equal, 0) };  
    \legend{Baseline 1 (F1), Baseline 2 (F1)} 
    \end{axis}  
    \end{tikzpicture}  
\caption{Performance of baseline models on the TIMELINE dataset in terms of F1-score.}
\label{fig:performance}
\end{figure}

\section{Evaluation Results and Ablation Study}
Table~\ref{tab:results1} and \ref{tab:results2} show the performance of the baseline models on the MATRES and TIMELINE datasets, respectively. \newcite{han2019deep} reported slightly different performance obtained by both models on MATRES. They discussed in their paper that they used three random seeds; however, since the value of these seeds were not made available, we have been unable to replicate the same results. 

As one can observe in Table \ref{tab:results2}, employing the second baseline model that adjusts for global constraints, leads to a performance improvement of 1.54 percentage points.
This is slightly higher than the improvement (0.26 percentage points) obtained by the second model on the MATRES corpus.
This is likely due to the higher number of globally consistent temporal relations in TIMELINE. 

In the subsections below, we discuss the main differences between the two corpora, MATRES and TIMELINE, and the impact of each key difference on the performance of the baseline models.  

\begin{table}[h]
\centering
\setlength\tabcolsep{5pt}
\begin{tabular}{|c|c|c|c|c|c|c|}
\hline
\multirow{2}{*}{}&\multicolumn{3}{|c|}{\textbf{{\small Baseline method 1}}}&\multicolumn{3}{|c|}{\textbf{{\small Baseline method 2}}}\\ \cline{2-7}
 & \textbf{{\small P}} & \textbf{{\small R}} & \textbf{{\small F1}} & \textbf{{\small P}} & \textbf{{\small R}} & \textbf{{\small F1}}  \\
\hline
\multirow{4}{*}{}{\small before} & {\small 70.07} & {\small 69.39} & {\small 69.73} & {\small 71.09} & {\small 71.92} & {\small 71.51}  \\
{\small after} & {\small 69.37} & {\small 71.53} & {\small 70.44} & {\small 71.09} & {\small 71.92} & {\small 71.51}  \\
{\small equal} & {\small 100}& {\small 8.33} & {\small 15.38} & {\small 0} & {\small 0} & {\small 0}  \\
\hline
{\small mi-F1}  & \multicolumn{3}{|c|}{{\small 69.74}} & \multicolumn{3}{|c|}{{\small 71.09}} \\  
\hline
 
\end{tabular}
\caption{\label{tab:ablation1}
Ablation study on Split 1A: the first split of the TIMELINE test set which contains relations involving verb-centred events.
}
\end{table}

\begin{table}[h]
\centering
\setlength\tabcolsep{5pt}
\begin{tabular}{|c|c|c|c|c|c|c|}
\hline
\multirow{2}{*}{}&\multicolumn{3}{|c|}{\textbf{{\small Baseline method 1}}}&\multicolumn{3}{|c|}{\textbf{{\small Baseline method 2}}}\\ \cline{2-7}
 & \textbf{{\small P}} & \textbf{{\small R}} & \textbf{{\small F1}} & \textbf{{\small P}} & \textbf{{\small R}} & \textbf{{\small F1}}  \\
\hline
\multirow{4}{*}{}{\small before} & {\small 60.86} & {\small 60.49} & {\small 60.68} & {\small 61.44} & {\small 62.96} & {\small 62.19}  \\
{\small after} & {\small 60.23} & {\small 63.58} & {\small 61.86} & {\small 61.44} & {\small 62.96} & {\small 62.19}  \\
{\small equal} & {\small 0}& {\small 0} & {\small 0} & {\small 0} & {\small 0} & {\small 0}  \\
\hline
{{\small mi-F1}}  & \multicolumn{3}{|c|}{{\small 60.54}} & \multicolumn{3}{|c|}{{\small 61.44}} \\  
\hline
 
\end{tabular}
\caption{\label{tab:ablation2}
Ablation study on Split 1B: the second split of the TIMELINE test set which contains relations involving non-verb-centred events.
}
\end{table}

\subsection{Inclusion of non-verb events}
News articles contain different events which, realistically, are not limited to events centred on verbs.
Thus, we investigated the impact of including non-verb events on model performance, particularly on the F1-score for the \texttt{before}, \texttt{after}, and \texttt{equal} temporal relations. Specifically, we sought to assess whether a model has learned relations involving non-verb events to the same extent that it has learned relations involving verb-centred events. 

To this end, we performed an ablation study by dividing the test set into two splits: (1) Split 1A contains samples with relations between verb events; and (2) Split 1B contains samples with relations where non-verb events are involved. We then evaluated the baseline models (trained and validated on the entire training/development sets) on each of the two splits. Unsurprisingly, we found that the performance on the first split (Table~\ref{tab:ablation1}) is higher than the performance on the second split (Table~\ref{tab:ablation2}). This indicates that the models are able to learn relations involving verb-centred events better than the relations with non-verb events during the training. This explains the higher performance of the models on the MATRES corpus given that it contains only verb-centred events. Moreover, one factor that may contribute to the reduced performance on the split that contains relations with non-verb events is that 70\% of the relations in the training and development splits involve only verb-centred events.

\begin{table}[h]
\centering
\setlength\tabcolsep{5pt}
\begin{tabular}{|c|c|c|c|c|c|c|}
\hline
\multirow{2}{*}{}&\multicolumn{3}{|c|}{\textbf{{\small Baseline method 1}}}&\multicolumn{3}{|c|}{\textbf{{\small Baseline method 2}}}\\ \cline{2-7}
 & \textbf{{\small P}} & \textbf{{\small R}} & \textbf{{\small F1}} & \textbf{{\small P}} & \textbf{{\small R}} & \textbf{{\small F1}}  \\
\hline
\multirow{4}{*}{}{\small before} & {\small 66.75} & {\small 66.08} & {\small 66.41} & {\small 67.48} & {\small 68.84} & {\small 68.15}  \\
{\small after} & {\small 65.94} & {\small 69.09} & {\small 67.48} & {\small 67.48} & {\small 68.84} & {\small 68.15}  \\
{\small equal} & {\small 100}& {\small 6.25} & {\small 11.76} & {\small 0} & {\small 0} & {\small 0}  \\
\hline
{{\small mi-F1}}  & \multicolumn{3}{|c|}{{\small 66.37}} & \multicolumn{3}{|c|}{{\small 67.48}} \\  
\hline
 
\end{tabular}
\caption{\label{tab:ablation3}
Ablation study on Split 2A: the first split of the TIMELINE test set containing temporal relations between events separated by at most 4 sentences.
}
\end{table}

\begin{table}[h]
\centering
\setlength\tabcolsep{5pt}
\begin{tabular}{|c|c|c|c|c|c|c|}
\hline
\multirow{2}{*}{}&\multicolumn{3}{|c|}{\textbf{{\small Baseline method 1}}}&\multicolumn{3}{|c|}{\textbf{{\small Baseline method 2}}}\\ \cline{2-7}
 & \textbf{{\small P}} & \textbf{{\small R}} & \textbf{{\small F1}} & \textbf{{\small P}} & \textbf{{\small R}} & \textbf{{\small F1}}  \\
\hline
\multirow{4}{*}{}{\small before} & {\small 69.45} & {\small 68.95} & {\small 69.20} & {\small 70.25} & {\small 70.75} & {\small 70.50}  \\
{\small after} & {\small 68.90} & {\small 70.39} & {\small 69.64} & {\small 70.25} & {\small 70.75} & {\small 70.50}  \\
{\small equal} & {\small 0}& {\small 0} & {\small 0} & {\small 0} & {\small 0} & {\small 0}  \\
\hline
{{\small mi-F1}}  & \multicolumn{3}{|c|}{{\small 69.17}} & \multicolumn{3}{|c|}{{\small 70.25}} \\  
\hline
 
\end{tabular}
\caption{\label{tab:ablation4}
Ablation study on Split 2B: the second split of the TIMELINE test set containing temporal relations between events separated by more than 4 sentences.
}
\end{table}

\subsection{Increasing the relation window}
Retrieving the temporal relation between any two events (regardless of how far they are from each other in a given news article) is an essential requirement for different tasks and domains. The three following cases are a good illustration of the importance of considering such types of relations.
\paragraph{Case 1:} In question answering (QA) tasks, to answer a specific time-based question, it is often necessary to retrieve the temporal relationship between an event in one of the first sentences and an event in the last few sentences of a given news article. It is impossible to retrieve this kind of long-distance relation in the previously published temporal information-annotated datasets since it is not tagged or cannot be retrieved using temporal reasoning (e.g., using transitive inference).
\paragraph{Case 2:} In the medical domain, to extract useful information (e.g., a timeline of medical events) from clinical notes and reports, it is important to identify temporal relations between events that are not in subsequent sentences.
\paragraph{Case 3:} Extracting a timeline of events from news articles allows decision-makers to conduct fine-grained analysis of these events; it is possible that events of interest are not in adjacent sentences. 
\\ \\
We set out to investigate the impact of increasing the relation window on model performance, particularly on the F1-score for the \texttt{before}, \texttt{after} and \texttt{equal} temporal relations. Specifically, we seek to determine if the models learned long-distance relations to the same extent as short-distance temporal relations.  

To this end, we conducted an ablation study based on the test set subdivided into two splits: (1) Split 2A contains examples with short-distance relations, i.e., relation window $<= 4$, and (2) Split 2B contains examples with long-distance relations, i.e., relation window $> 4$. We set the threshold to 4 considering that the average relation window in our corpus is 9. If we split the set of relations in the corpus in this way, the short-distance relations involve events with 0 to 4 sentences between them; the long-distance ones involve events with more than 4 sentences between them. 

The trained baseline models were then evaluated on each of the two splits. Interestingly, the performance on the second split (Table~\ref{tab:ablation4}) is higher than on the first split (Table~\ref{tab:ablation3}). This demonstrates that the models have learned long-distance relations better than short-distance temporal relations during the training process. A contributing factor to this is the fact that Split 2A has a slightly larger percentage of non-verb events, which we now know are more difficult for the models to learn (26.11\% of the relations), compared with Split 2B (21.51\% of the relations).

\begin{table}[ht]
\centering
\setlength\tabcolsep{5pt}
\begin{tabular}{|c|c|c|c|c|c|c|}
\hline
{\small Datasets} & {\small Splits} & \begin{tabular}{@{}c@{}} {\small \# of }  \\
{\small possible}  \\{\small pairs}  \end{tabular} & \begin{tabular}{@{}c@{}} {\small \# of non-vague} \\ {\small pairs annotated}  \end{tabular} & {\small \%}  \\
\hline
\multirow{4}{*}{\textbf{{\small MATRES}}} & {\small Train}  & {\small 14219} & {\small 943} & {\small 6.63}    \\
 & {\small Dev}  & {\small 916} & {\small 198} & {\small 21.61}   \\
  & {\small Test}  & {\small 1433} & {\small 272} & {\small 18.98}   \\
  & {\small Total}  & {\small 16568} & {\small 1413} & \textbf{{\small 8.52}}    \\ 
\hline
\multirow{4}{*}{\textbf{{\small TIMELINE}}} & {\small Train}  & {\small 5215} & {\small 2384} & {\small 45.71}   \\
 & {\small Dev}  & {\small 1313} & {\small 284} & {\small 21.63}   \\
  & {\small Test}  & {\small 2028} & {\small 685} & {\small 33.78}   \\
  & {\small Total}  & {\small 8556} & {\small 3353} & \textbf{{\small 39.19}}   \\ 
\hline

\end{tabular}
\caption{\label{tab:analysis1}
Number and proportion of annotated temporal relations in MATRES and TIMELINE.
}
\end{table} 

\begin{table}[ht]
\centering
\setlength\tabcolsep{5pt}
\begin{tabular}{|c|c|c|c|c|c|c|}
\hline
{\small Datasets} & {\small Splits} & \begin{tabular}{@{}c@{}} {\small \# of }  \\
{\small possible}  \\{\small pairs}  \end{tabular} & \begin{tabular}{@{}c@{}} {\small \# of non-vague}  \\ {\small pairs classified} \\ {\small correctly} \end{tabular} & {\small \%}  \\
\hline
\multirow{4}{*}{\textbf{{\small MATRES}}} & {\small Train}  & {\small 14219} & {\small -} & {\small -}   \\
 & {\small Dev}  & {\small 916}  & {\small -} & {\small -}  \\
  & {\small Test}  & {\small 1433} & {\small 235} & \textbf{{\small 16.39}}  \\
  & {\small Total}  & {\small 16568}  & {\small -} & {\small -}   \\ 
\hline
\multirow{4}{*}{\textbf{{\small TIMELINE}}} & {\small Train}  & {\small 5215} & {\small -} & {\small -}   \\
 & {\small Dev}  & {\small 1313} & {\small -} & {\small -}  \\
  & {\small Test}  & {\small 2028}  & {\small 473} & \textbf{{\small 23.32}}  \\
  & {\small Total}  & {\small 8556} & {\small -} & {\small -}   \\ 
\hline

\end{tabular}
\caption{\label{tab:analysis2}
Number and proportion of temporal relations in MATRES and TIMELINE that were automatically extracted by the better baseline model (the second model).
}
\end{table} 
\subsection{Extracting more relations}
In an earlier section, we have shown that models find it more challenging to learn relations involving non-verb events, compared with verb-centred events. Despite the lower performance of the two baseline models on our proposed dataset, the models are able to extract more temporal relations than in MATRES. 

As shown in Table~\ref{tab:analysis1}, in MATRES, only 8.52\% of the possible relations were annotated as non-vague; meanwhile, 39.19\% of the possible relations were labelled as non-vague in TIMELINE. 

In Table~\ref{tab:analysis2}, we show that the second baseline model extracted only 16.39\% of the possible relations in the test set of the MATRES dataset.
In our proposed dataset, TIMELINE, the model was able to extract 23.32\% of the possible relations in the test set.

\section{Reasoning Behind this Annotation in the LLMs Era}
We believe that despite the advent of large language models (LLMs), this kind of fine-grained annotation is still necessary to support the development of supervised models.
We argue that the temporal relation extraction performance of an LLM such as ChatGPT, for example, is not comparable in relation to that of supervised models.
Firstly, \newcite{yuan2023zero} investigated ChatGPT's capability in zero-shot temporal relation extraction and showed that ChatGPT's performance is lower by up to 30\% in terms of F1-score compared to supervised methods.
Furthermore, we investigated the extent to which ChatGPT can extract temporal relations by prompting it using the zero-shot prompt proposed by \newcite{yuan2023zero} to identify temporal relations between events in the TIMELINE test set. 
Overall, ChatGPT obtained precision, recall and F1-scores of 31.11\%, 35.67\% and 33.24\%, respectively.
These are substantially lower than those of the second baseline method, which obtained 69.05\% for precision, 69.05\% for recall and 69.05\% for F1-score.

\section{Potential Applications}
Our annotation scheme and dataset hold promise for various practical uses. Extracting temporal relations from news articles can support information extraction applications such as automatic timeline extraction and question answering (QA). Moreover, considering that the focus of the dataset is on events on the main axis (i.e., events in the main storyline), this work can potentially support narrative extraction applications such as the analysis of events related to financial markets and event monitoring, e.g., in the context of disaster management \citep{norambuena2023survey}.

\section{Conclusion}
In this paper, we present a new corpus, TIMELINE, which was annotated following a novel annotation scheme whereby non-verb-centred events are included, as well as long-distance temporal relations between events.
The corpus was used in training and evaluating two baseline temporal relation extraction models.
Based on our evaluation results, we assessed the impact of increasing the relation window and including non-verb-centred events on model performance. In addition, we demonstrated how our annotation scheme can support the development of models that can extract more relations in comparison with earlier datasets. In the future, we aim to increase the size of the dataset and employ it in a timeline generation task. 

\section*{Limitations}
This temporal relation research focussed on a specific type of publication, namely, newspapers articles published on a daily basis. As a result, we did not consider other types of publications which are published weekly or monthly. The primary motivation for this consideration is to use the narrative container concept \citep{pustejovsky2011increasing}, which has helped significantly to increase our annotation accuracy. Also, as we mentioned previously, we considered only events that can be anchored onto a timeline and that belong to the main axis (storyline).

\bibliography{anthology,custom}
\bibliographystyle{acl_natbib}

\appendix

\section*{Appendix}
\label{sec:appendix}
\subsection*{Annotation Guidelines}
\label{sec:annotation_guidelines}
\paragraph{Step 1: Event annotation.}
All events according to the TimeML guidelines  \citep{sauri2006timeml} will be tagged, except for the following:
\begin{enumerate}
\item Cancelled or negated events will not be tagged; for example, \emph{``He failed to \textbf{find} buyers''}, \emph{``They don't \textbf{want} to play with us''}, or \emph{``She cancelled the \textbf{meeting}''}. Moreover, uncertain events will not be annotated, e.g., \emph{``We may \textbf{go}.''}
\item Inspired by the TimeML guidelines, the following events will not be tagged: (1) generics (abstract and non-specific events), e.g., \emph{``Fruit \textbf{contains} water.''}, \emph{``Lions \textbf{hunt} Zebra.''}; (2) static events, e.g., \emph{``New York \textbf{is} on the east coast.''}
\item Hypothetical/conditioned events will not be annotated. For example, \emph{``If I'm \textbf{elected} as president, I will \textbf{cut} income tax for everyone.''}  
\item Inspired by the annotation scheme followed by \citep{minard2015semeval}, adjectives express the property or attribute of an entity and anchoring them in time is not simple. Thus, adjectives will not be tagged. 
\item Events after modal verbs will not be tagged. For example, \emph{``We have to \textbf{leave}.''}, or \emph{``You must be sending the email by the end of the day.''}
\item Intended events will not be tagged. They express intentions or things that are meant to happen or occur and signified by words such as \emph{``plan''}, \emph{``aim''}, \emph{``intend''} and \emph{``hope''}.
\end{enumerate}

\begin{figure}
 \includegraphics[width=7.5cm, height=3cm]{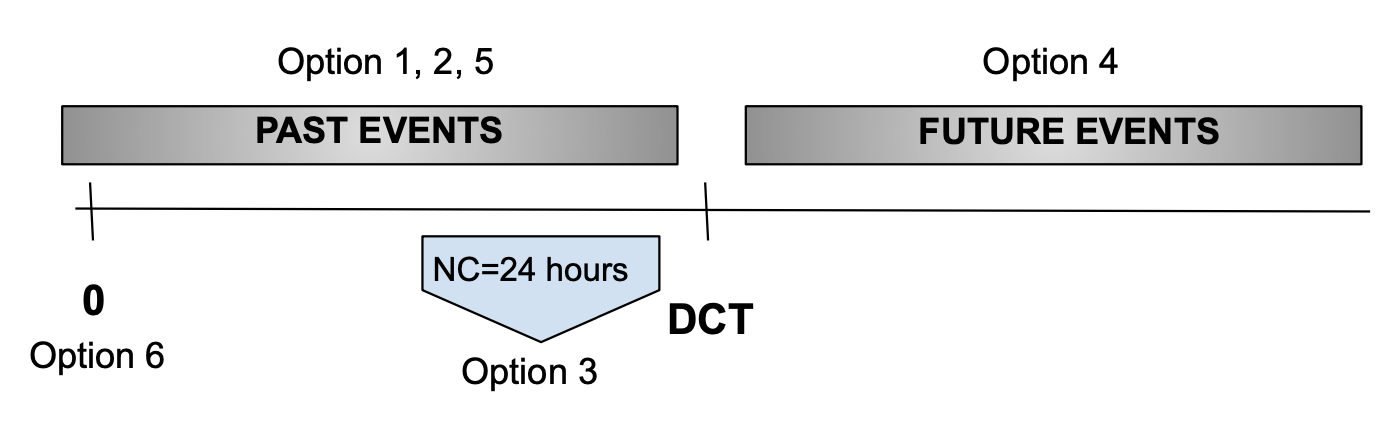}
 \caption{Timeline modelling for events}
 \label{fig:Timeline}
\end{figure}

\paragraph{Step 2: Time anchor annotation.}
The annotators were asked to enter the time anchor for each event by choosing one out of six options: 
\begin{itemize}
\item \textbf{Option 1}: If the text explicitly mentions the time of the event (e.g., \emph{``Feb 1, 2021''}), the annotator should enter that date as a time anchor for the event.
If the text does not mention the exact date but uses temporal expressions that are relative to the document creation time (DCT), e.g., \emph{``today''}, \emph{``last Friday''}, the annotator should use the calendar to enter the date in relation to the DCT. 
\item \textbf{Option 2}: If the text implicitly mentions the event's time (e.g., \emph{``last August''}), the annotator should enter the date as a fuzzy date (e.g., \emph{``2020-08-XX''}). Alternatively, if the text mentions that the event happened last year, the annotator should enter e.g., \emph{``2020-XX-XX''}. 
\item \textbf{Option 3}: If the event has no temporal information, but it is clear from the text that the event happened around the document creation time (DCT), the date should be set to the default narrative container (NC) value for newspaper publications which is one day before the DCT. 
\item \textbf{Option 4}: If the event happens in the future, the default date will be one day after the DCT. Alternatively, if it is mentioned in the text that the event will happen sometime relative to the DCT, e.g.,  \emph{``next Friday''}, the annotator can enter that day's date.
\item \textbf{Option 5}: If the event happened in the past but the time is not mentioned in the text explicitly, the annotators can use any background or external knowledge to provide an accurate time anchor.
\item \textbf{Option 6}: If the annotator understands from the text that the event did not happen around the document creation time, and the text does not provide any hints on when the event happened, the date should be entered as \emph{``XXXX-XX-XX''}. Figure  \ref{fig:Timeline} shows how the events are represented in a timeline.
\end{itemize}

\paragraph{Step 3: Answer a set of questions for each annotated event.}

\begin{itemize}
\item \textbf{Question 1:} To annotate the relation between events that are the same (event coreference) with an \texttt{equal} label.

Q1: \emph{Does the event refer to another event in the document?} (Q1.a: Yes/No, Q1.b: event ID).

\item \textbf{Question 2:} To annotate temporal relations with an \texttt{equal} label. 

Q2: \emph{Did the event start or happen at the same time when another event in the same sentence happened?} (Q2.a: Yes/No, Q2.b: event ID).

\item \textbf{Questions 3, 4 and 5:}
To increase informativeness, i.e., to increase the number of non-vague relations.

Q3: \emph{Did the event happen on the same day as another event in the same sentence? If so, did the event happen at a different time compared with the other event?} (Q3.a: Yes/No, Q3.b: before/after, Q3.c: event ID)

Q4: \emph{Is this event with an unknown date (Option 6)? If so, did it happen before or after another event in the same sentence?} (Q4.a: Yes/No, Q4.b: before/after, Q4.c: event ID)

Q5: \emph{Were this event and another event in the same sentence given the same implicit time? If so, did this event happen before/after the other one?} (Q5.a: Yes/No, Q5.b: before/after, Q5.c: event ID)

\item \textbf{Question 6:} To annotate the relation between events that happened around the DCT but were given different time anchors, as \texttt{vague}.

Q6: \emph{Did the event happen around the document creation time (e.g., within 24 hours)? (Yes/No)}

For instance, consider the two events in the following sentences. Sentence 1: \textit{``The pound gained almost 6 per cent against the dollar in July, \textbf{approaching} \$1.32 at one point yesterday before settling in evening trading at \$1.31, up 0.04 per cent for the day and 5.7 per cent for the month.''}. The event \emph{``approaching''} happened \emph{``yesterday''}. The temporal information is mentioned explicitly in the text for this event. Sentence 2: \textit{``Bank of America Merrill Lynch strategists \textbf{said} the rest of 2020 could still see weakness for the pound as the period of August to December historically contains four negative months for sterling.''} The event \emph{``said''} happened possibly one day before the publication date (based on what the reader of the news article could infer according to the narrative container concept). However, it is not clear from the text which of the two events \emph{``approaching''} or \emph{``said''}  happened first. Therefore, if the annotator answered the question with \texttt{Yes} for both events, our temporal relation generator will assign the relation label \texttt{vague} to these events to ensure accuracy.

\item \textbf{Question 7:} To annotate a relation between events that are happening in the future but were given different time anchors, as a \texttt{vague} relation.

Q7: \emph{Is the event happening in the future? (Yes/No)}

For instance, sometimes in the text, it is mentioned that some event (\textit{Event 1}) will happen in the future without any time anchor; for another event (\textit{Event 2}), the text says that it will occur at a specific time (e.g., \emph{``next month''}). However, it might be unclear from the text which event will happen first. Therefore, if the annotator answered the question with \texttt{Yes} for both events, our temporal relation generator will assign the relation \texttt{vague} to these events.

\end{itemize}

\paragraph{Step 4: Temporal relation annotation.}
The temporal relations are annotated automatically based on Algorithm~\ref{algorithm1}.

\begin{algorithm*}[hbt!]
\caption{Temporal Relation Generation Method}
\begin{algorithmic} 
\STATE \textbf{M} = all possible event pairs($ai,bi$) in the document 
\FOR{i in M}
\small{\IF{$((time(ai)==time(bi))\ \AND\ (((Q1.a(ai)==\texttt{Yes})\ \AND\ (Q1.b(ai)==bi))\ \OR\ ((Q2.a(ai)==\texttt{Yes})\ \AND\ (Q2.b(ai)==bi)))) $} 
\STATE $Label = \texttt{equal} $
\ELSIF{$(((time(ai)<time(bi)) \ \AND\ (((Q6(ai )\neq \texttt{Yes})\ \OR\ (Q6(bi) \neq \texttt{Yes}))\ \AND\ ((Q7(ai )\neq \texttt{Yes})\ \OR\ (Q7(bi) \neq \texttt{Yes})))) \ \OR\ ((time(ai)==time(bi)) \ \AND\ (Q3.a(ai)==\texttt{Yes}) \ \AND\ (Q3.b(ai)==\texttt{before})\ \AND\ (Q3.c(ai)==bi))\ \OR\ ((Q4.a(ai)==\texttt{Yes})\ \AND\ (Q4.b(ai)==\texttt{before})\ \AND\ (Q4.c(ai)==bi))\ \OR\ ((Q5.a(ai)==\texttt{Yes})\ \AND\ (Q5.b(ai)==\texttt{before})\ \AND\ (Q5.c(ai)==bi))) $} 
\STATE $Label = \texttt{before} $
\ELSIF{$(((time(ai)>time(bi)) \ \AND\ (((Q6(ai)\neq \texttt{Yes}) \ \OR\ (Q6(bi) \neq \texttt{Yes})) \ \AND\ ((Q7(ai)\neq \texttt{Yes}) \ \OR\ (Q7(bi) \neq \texttt{Yes}))))\ \OR\ ((time(ai)==time(bi))\ \AND\ (Q3.a(ai)==\texttt{Yes})\ \AND\ (Q3.b(ai)==\texttt{after})\ \AND\ (Q3.c(ai)==bi)) \ \OR\ ((Q4.a(ai)==\texttt{Yes}) \ \AND\ (Q4.b(ai)==\texttt{after}) \ \AND\ (Q4.c(ai)==bi))\ \OR\ ((Q5.a(ai)==\texttt{Yes})\ \AND\ (Q5.b(ai)==\texttt{after})\ \AND\ (Q5.c(ai)==bi)))  $} 
\STATE $Label = \texttt{after} $
\ELSE
\STATE $Label = \texttt{vague} $
\ENDIF}
\ENDFOR
\end{algorithmic}
\label{algorithm1}
\end{algorithm*}

\subsection*{Special Cases}
Below are two cases encountered during the annotation process that we needed to make the annotators aware of.
\begin{itemize}
\item \textbf{Event Coreference:} When we have more than two events referring to the same thing, the relations that involve these events have to be annotated manually after Step 4.

\item \textbf{Subsequent events only:} In Q1, the annotators should verify that the event ID is associated with a subsequent event mentioned in the same or any following sentence. In Q2-Q5, the annotators should ensure that the event ID is associated with a subsequent event mentioned in the same sentence.

\end{itemize}
\end{document}